\documentclass[journal]{IEEEtran}
\usepackage[T1]{fontenc}
\usepackage{color}
\usepackage{flushend}
\usepackage{amsmath}
\usepackage[ruled,linesnumbered]{algorithm2e}
\usepackage{algpseudocode} 
\usepackage{makecell}
\usepackage{booktabs}
\usepackage{tabu}
\usepackage[cmintegrals]{newtxmath}
\usepackage{bm}
\usepackage{indentfirst}
         
\usepackage{amsmath,amssymb}
\usepackage{array}
\usepackage{mathrsfs}
\usepackage{cite}
\usepackage{tikz}
\usepackage{color}
\usepackage{amssymb}
\usepackage{url}
\usepackage{hyperref}
\usepackage{todonotes}
\usepackage{color}
\usepackage{bbding}
\usepackage{pifont}

\begin{document}

\title{Digital and Physical Face Attacks: Reviewing and One Step Further}

\author{Chenqi~Kong, 
Shiqi~Wang,~\IEEEmembership{Senior Member,~IEEE},
and Haoliang~Li,~\IEEEmembership{Member,~IEEE}
\thanks{C. Kong, and S. Wang are with the Department of Computer Science, City University of Hong Kong, Hong Kong, China. (email: cqkong2-c@my.cityu.edu.hk; shiqwang@cityu.edu.hk).}% <-this % stops a space
\thanks{H. Li is with the Department of Electrical Engineering, City University of Hong Kong, Hong Kong, China. (email: haoliang.li@cityu.edu.hk).}
% \thanks{A. Rocha is with the Artificial Intelligence Lab. (\texttt{Recod.ai}) at the University of Campinas, Campinas 13084-851, Brazil (e-mail: anderson.rocha@ic.unicamp.br), URL: \url{http://recod.ai}}
% \thanks{H. Li is the corresponding author.}
}

% The paper headers
\markboth{}%
{Shell \MakeLowercase{\textit{et al.}}: Bare Demo of IEEEtran.cls for IEEE Communications Society Journals}

\maketitle

\begin{abstract}
With the rapid progress over the past five years, face authentication has become the most pervasive biometric recognition method. Thanks to the high-accuracy recognition performance and user-friendly usage, automatic face recognition (AFR) has exploded into a plethora of practical applications over device unlocking, checking-in, and financial payment. In spite of the tremendous success of face authentication, a variety of face presentation attacks (FPA), such as print attacks, replay attacks, and 3D mask attacks, have raised pressing mistrust concerns. Even worse, as attack techniques are getting more and more powerful and smart, FPA is becoming increasingly realistic and advanced. Besides physical face attacks, face videos/images are vulnerable to a wide variety of digital attack techniques launched by malicious hackers, causing potential menace to the public at large. Due to the unrestricted access to enormous digital face images/videos and disclosed easy-to-use face manipulation tools circulating on the internet, non-expert attackers without any prior professional skills are able to readily create sophisticated fake faces, leading to numerous dangerous applications such as  financial fraud, impersonation, and identity theft. Nowadays, face information has become the dominant biometric trait of a person and unique non-verbal but powerful FaceID. How to safeguard personal face information against both physical and digital attacks is of great importance. This survey aims to build the integrity of face forensics by providing thorough analyses of existing literature and highlighting the issues requiring further attention. In this paper, we first comprehensively survey both physical and digital face attack types and datasets. Then, we review the latest and most advanced progress on existing counter-attack methodologies and highlight their current limits. Moreover, we outline possible future research directions for existing and upcoming challenges in the face forensics community. Finally, the necessity of joint physical and digital face attack detection has been discussed, which has never been studied in previous surveys. 

\end{abstract}


\IEEEpeerreviewmaketitle
\section{Introduction}
Significant progress on face recognition techniques has been made since the advent of Apple's highly touted FaceID and the follow-up face authentication works. Face recognition systems have pervaded into billions of people's daily lives over various applications such as device unlocking, log-in, and e-banking. Consequently, face information nowadays has become the dominant biometric trait of a person, a unique FaceID, and a vehicle itself of non-verbal but powerful messages \cite{verdoliva2020media}. With the rapid proliferation of face multimedia content circulating on social media platforms, unrestricted access to digital media content has posed high level of risks over privacy leakage, identity theft, and financial fraud. In spite of the achieved tremendous success on face authentications, potential malicious face attacks, including digital and physical attacks, have raised pressing security concerns to the public at large.

As shown in Fig.~\ref{digital_fake}, digital face attacks can be basically classified into four categories: (1). identity swap; (2). face reenactment; (3). attribute manipulation; and (4). entire face synthesis \cite{dang2020detection}. Identity swap \cite{hm16_20, dfcode, bitouk2008face} is actually not a new problem. The first ever work on identity swap dates back to 1860, where Abraham Lincoln's head is stitched up with the body of southern politician John Calhoun \cite{Lincoln}. Heading to the era of artificial intelligence and deep learning, deepfake techniques, employing  powerful various generative models, are able to create sophisticated fake faces with the target identity. Face reenactment  ($a.k.a.$ expression edition) \cite{thies2016face2face, thies2019deferred} aims to transfer the source person's facial expression to the target one. Face2Face \cite{thies2016face2face} and NeuralTextures \cite{thies2019deferred} are two of the most prominent facial expression editing techniques. Moreover, attribute manipulation \cite{choi2018stargan, he2019attgan} empowered by numerous image translation methods \cite{isola2017image, zhu2017unpaired, zhu2017toward} attempts to edit face attributes such as hair, glasses, and skin color in face images/videos. 
Thanks to the recent advances of various generative models \cite{goodfellow2014generative, van2016pixel, van2016conditional, kingma2013auto, kingma2019introduction}, entire face synthesis \cite{karras2019style, karras2018progressive} can generate face pictures whose identity does not exist with a high level of realism. Generally speaking, attribute manipulation and entire face synthesis techniques tend to bring positive impacts to human lives, while identity swap and face reenactment could easily cause disconcerting security problems. For this reason, this survey mainly focuses on identity swap and face reenactment. 

\begin{figure*}[h]
\centering
\includegraphics[scale=0.62]{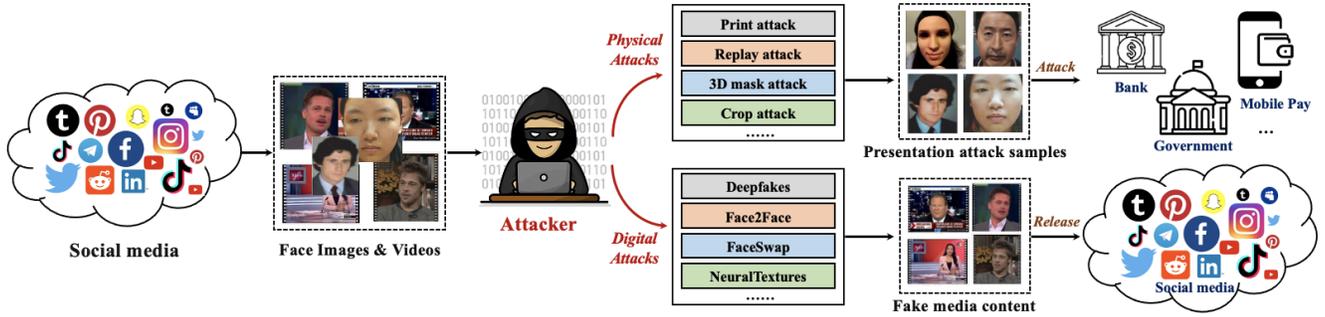}
\caption{Overview of physical and digital face attacks. }
\label{pipeline}
\end{figure*}

\begin{table*}
  \caption{Comparisons with prior survey papers}
  \label{prior_surveys}
  \centering
  \renewcommand\arraystretch{1.15}
  \scalebox{1.0}{\begin{tabular}{|c|c|c|c|c|c|c|}
    \hline
    Prior Surveys & Timelines & Ref. Scale & \#Dataset & Physical Attack & Digital Attack & Unified Attack \\
    \hline
    Souza et al.~\cite{souza2018far} & 2018 & 98 & 9 & \checkmark & - & - \\
    \hline
    Raheem et al.~\cite{raheem2019insight} & 2019 & 90 & 14 & \checkmark & - & - \\
    \hline
    Pereira et al.~\cite{pereira2020rise} & 2019 & 57 & 7 & \checkmark & - & - \\
    \hline
    Jia et al.~\cite{jia2020survey} & 2020 & 74 & 10 & \checkmark & - & - \\
    \hline
    Safaa et al.~\cite{safaa2020deep} & 2020 & 127 & 8 & \checkmark & - & - \\
    \hline
    Kotwal et al.~\cite{kotwal2019multispectral} & 2020 & 42 & 12 & \checkmark & - & - \\
    \hline
    Yu et al.~\cite{yu2021deep} & 2022 & 252 & 36 & \checkmark & - & - \\
    \hline
    \hline
    Nguyen et al.~\cite{nguyen2019deep} & 2019 & 106 & 3 & - & \checkmark & - \\
    \hline
    Verdoliva~\cite{verdoliva2020media} & 2020 & 274 & 9 & - & \checkmark & - \\
    \hline
    Lyu~\cite{lyu2020deepfake} & 2020 & 34 & 5 & - & \checkmark & - \\
    \hline
    Tolosana et al.~\cite{tolosana2020deepfakes} & 2020 & 200 & 7 & - & \checkmark & - \\
    \hline
    Mirsky and Lee~\cite{mirsky2021creation} & 2021 & 192 & 3 & - & \checkmark & - \\
    \hline
    \hline
    Ours & 2022 & 253 & 32 & \checkmark & \checkmark & \checkmark \\
    \hline
\end{tabular}}
\end{table*}

\begin{figure}[h]
\centering
\includegraphics[scale=0.45]{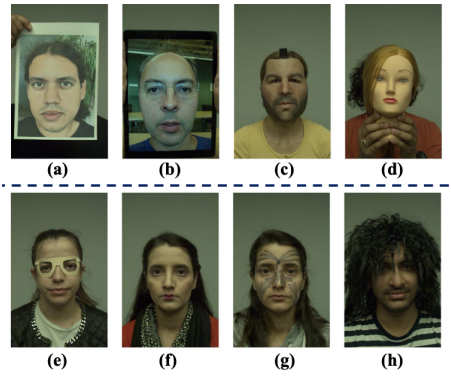}
\caption{Typical face presentation attack (FPA) examples \cite{heusch2020deep}. The top row shows the impersonation attacks: (a). print, (b). replay, (c). 3D mask, and (d). mannequin. The bottom row presents the obfuscation attacks: (e). glasses, (f). makeup, (g). tattoo, and (h). wig.}
\label{fpa_samples}
\end{figure}

Based on the attack techniques and intents, physical face attacks ($a.k.a.$ Face presentation attacks (FPA)) can be broadly categorized into two classifications: impersonation and obfuscation. As shown in Fig.~\ref{fpa_samples} (a)-(d), impersonation attacks include typical print attack, replay attack, and 3D mask attack, where attackers impersonate the target identity by covering the whole face region to fool face recognition systems. Generally speaking, 2D print and replay attacks can be easily launched by non-expert persons. In turn, 3D mask attacks, including silicon masks, resin masks, plastic masks, and mannequins, always demand advanced fabrication systems to capture the target person’s 3D facial information, which requires great efforts and costs  \cite{patel2016secure}. 
On the other hand, more advanced FPA types have been subsequently proposed, such as makeup attack, tattoo attack, funny glasses attack, and wig attack, as shown in Fig.~\ref{fpa_samples} (e)-(h). We cast these attacks as the obfuscation attack, where attackers partially obfuscate the face region to hide the attacker's identity. Compared with the impersonation FPA, the latter is more realistic and challenging to detect.  

\begin{figure*}[h]
\centering
\includegraphics[scale=0.31]{dataset.pdf}
\caption{Four digital face attack types and corresponding forgery regions. White pixels indicate the forged region.}
\label{digital_fake}
\end{figure*}

Malicious hackers can handily download the media content circulating on the internet and launch two types of face attack: physical and digital face attacks. Fig.~\ref{pipeline} illustrates the general pipeline. On the side of physical attacks, numerous face presentation attacks, such as 2D print/replay attack, 3D mask attack and makeup attack, can be easily launched to hack the face authentication systems over various application scenarios. On the other side, the attacker can also employ off-the-shelf APPs ($e.g.$, ZAO \cite{zao}, Facebrity \cite{Facebrity}, and Reface \cite{Reface}) or disclosed face manipulation algorithms ($e.g.$, Deepfakes \cite{hm16_20}) to edit face content fueled with targeted disinformation or misinformation. The generated fake content can be released or disseminated to the social network platforms, causing detrimental mistrust issues. Even worse, as the attack methodologies are getting increasingly advanced, the produced fake faces are becoming more and more sophisticated. Powerful as the attack technique is, there is a thin line between bonafide and fake faces that can be hardly distinguished by human naked eyes, and it is easy to cross over. To that end, the abuse of either physical or digital face attacks will certainly lead to the tendency of reducing the trust of digital media content and raising tangible concerns, in the long run. 

\begin{figure}[h]
\centering
\includegraphics[scale=0.35]{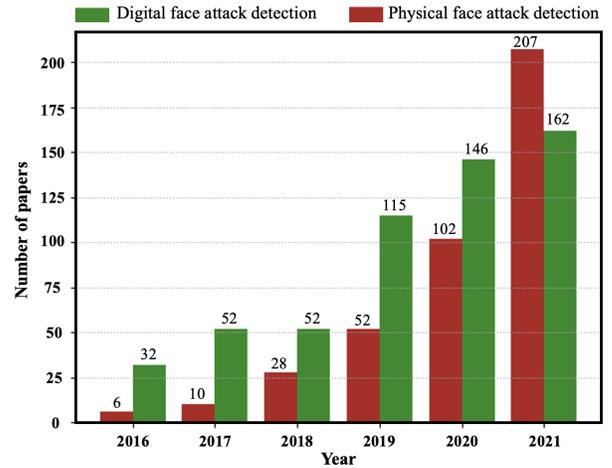}
\caption{The growing trend in the number of papers in the digital face attack detection and physical face attack detection fields \cite{juefei2022countering, yu2021deep}.}
\label{number}
\end{figure}

To counter various malicious physical face attacks and safeguard face recognition systems, numerous traditional methods have been first proposed. These methods mine informative artifacts via extracting hand-crafted features such as histograms, gradients, and texture~\cite{maatta2011face,de2012lbp, komulainen2013context, patel2016secure, boulkenafet2016face, peixoto2011face}. With the advent of deep learning, the accuracy of learning-based PAD methodologies~\cite{yu2021revisiting, jia2020single, li2018unsupervised, zhang2020face, cai2020drl, shao2020regularized, yu2020multi, liu2018learning} significantly boosts. To overcome the overfitting problem of the data-driven models, some methods seek to employ auxiliary modality information, such as remote physiological signals (rPPG)~\cite{li2016generalized, liu2018learning, lin2019face, yu2019remote}, pseudo depth maps~\cite{liu2018learning, atoum2017face, yu2020searching, wang2020deep, yu2020fas, zhang2020face}, Near-infrared (NIR) maps~\cite{yu2020multi, sun2016context, liu2021data, liu2021face}, and Thermal maps~\cite{seo2019face, mallat2021indirect}. On the other hand, great efforts have been dedicated to tackling the digital face attack problems over the past five years. Most existing face forgery detection algorithms are AI- or deep-learning based~\cite{chollet2017xception, tan2019efficientnet, qian2020thinking, masi2020two, zhao2021multi, kong2022detect, miao2022hierarchical}. Apart from some models focusing on distinguishing input face images/videos between real and fake, some recent works~\cite{dang2020detection, kong2022detect, wang2022lisiam, yu2022improving, huang2022fakelocator} propose to localize forged regions for fake face appearances. Unsurprisingly, most learning-based detectors suffer significant performance drops when deployed to unforeseen datasets or attack types. As such, \cite{li2020face, liu2021spatial, haliassos2021lips, luo2021generalizing, shiohara2022detecting, zhao2021learning} designed more generalized face forgery detection models that mine more inherent clues of fake media and mitigate the severe domain gaps. Fig.~\ref{number} shows the explosive increase of published literature numbers for both digital and physical face attack detection in recent years. It can be seen that significant efforts have been devoted to the face forensics community, making it an active research area.

% Generalization~\cite{li2020face, liu2021spatial, haliassos2021lips, luo2021generalizing, shiohara2022detecting, zhao2021learning}

Compared with previous surveys, we note that this survey is unique and superior in the following three aspects:

\begin{itemize}
    \item[$\bullet$] We, for the first time to the best of our knowledge, aggregate and examine the literature on both digital and physical face attacks into one survey. We outline that the unified face attack detection would be a promising research area in the face forensics community. 
    \item[$\bullet$] As shown in Table. \ref{prior_surveys}, this survey provides, by far, the latest and most comprehensive overview of the face forensics literature (>250 research papers) over attack types, datasets, and detection methodologies. 
    \item[$\bullet$] This survey poses severe security, privacy, and explainability issues that have been largely understudied in the existing literature. Based on the outlined issues, we further suggest future research orientations to facilitate the development of this community. 
\end{itemize}

This survey starts with reviewing the physical face attacks and digital face attacks in Sec. II and Sec. III, respectively. We briefly discuss the importance of the problem and concretely review related literature regarding attack types, datasets, and detection methodologies. Then we analyze the existing security issues and suggest possible future research directions to facilitate the development of the face forensics community. Sec. IV innovatively investigates the unifying detection works against face spoofing and face forgery. We also thoroughly analyze and discuss the motivations, benefits, and future research of the unified face attack defense. Finally, we draw the conclusion in Sec. V.
% detrimental outcomes, in line with
% faces play a central role in human communication , as the face of a person can emphasize a message or it can even convey a message in its own right \cite{frith2009role}. 

\section{Physical face attacks}
\begin{figure}[h]
\centering
\includegraphics[scale=0.40]{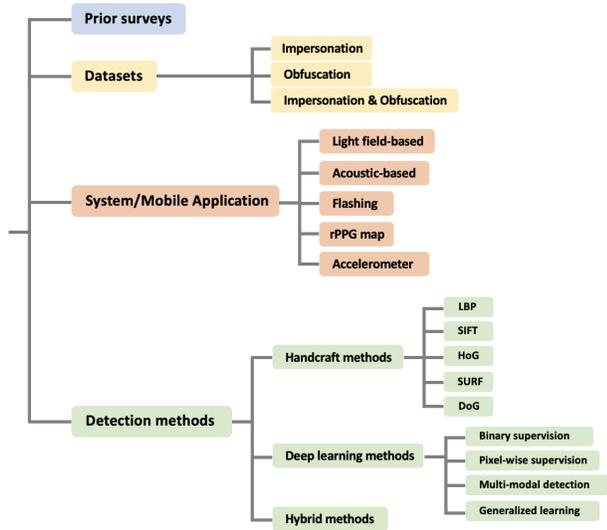}
\caption{Tree diagram of physical face attack paper structure.}
\label{fas_tree_diagram}
\end{figure}

As automatic face recognition (AFR) systems have been prevalently deployed in a wide variety of applications, face presentation attack detection (PAD) has attracted extensive attention from both industry and academia. It is of utmost necessity to safeguard AFR against malicious physical face attacks. In this section, we provide a comprehensive review on the literature of physical face attacks. The main literature structure is illustrated in Fig.~\ref{fas_tree_diagram}.

\subsection{Importance of the problem}
Physical face attacks, also known as face presentation attacks (FPAs), can be deployed either as an obfuscation attack or as an impersonation attack, where the former attempts to hide one's identity and the latter aims at impersonating the target person. Past decades have witnessed the rapid proliferation of face authentication systems, and they have exploded into various practical applications ranging from log-in, financial payment, check-in, etc. FPA is getting increasingly notorious because it can easily bypass the face authentication system. For example, Apple's FaceID was hacked by a 3D mask FPA~\cite{FaceID_attack} in 2017 and caused disturbing security concerns. With the rapid development of attack methods and fabrication techniques in this era, FPAs tend to be more and more sophisticated and challenging. As such, it is of great importance to design highly accurate and secure presentation attack detection (PAD) models to safeguard face recognition systems against FPAs.

\subsection{Face presentation attacks and datasets}
 Face presentation attacks can be generally divided into two categories: obfuscation attacks and impersonation attacks. As shown in Fig.~\ref{fpa_samples} (e)-(h), obfuscation attacks such as glasses, makeup, tattoo, and wig attempt to hide someone's identity. On the other hand, impersonation attacks (Fig.~\ref{fpa_samples} (a)-(d)) attempt to mimic the target person's identity by copying the target person's face to specific mediums, such as paper, screen, and 3D mask. Over the past fifteen years, substantial efforts have been devoted to building face anti-spoofing (FAS) datasets for facilitating the algorithm design of presentation attack detection. In Table.~\ref{FAS_databases}, we comprehensively summarize the existing face presentation attack databases in terms of modality, quantity, spoof medium, and acquisition device. To fit the uncontrollable environmental variables ($e.g.$, illumination, scene, acquisition device, spoof medium, etc.) in practical scenarios, face anti-spoofing (FAS) databases tend to be increasingly diverse. On the other hand, scale is another pivotal factor of FAS databases, as most deep learning-based methods demand large-scale training data to guarantee high-level PAD performance when deployed in real-world applications. Besides the RGB vision modality, more modalities such as depth map, near-infrared (NIR), thermal map, flashing, and acoustic have been gradually incorporated in lastly released databases. The additional modalities can serve as auxiliary information to improve the generalization capability of PAD models. Due to the two-player nature between FPA and PAD, novel presentation attacks with higher quality will be constantly proposed with the development of smart attack algorithms and advanced fabrication techniques. As such, it is unsurprising that attack types tend to be more and more diverse in newly published databases. 

% All the factors mentioned above contribute significantly to the face forensics community.   

\begin{table*}
  \caption{Summary of face presentation attack databases}
  \label{FAS_databases}
  \centering
  \renewcommand\arraystretch{1.15}
  \scalebox{0.95}{\begin{tabular}{|c|c|c|c|c|c|}
    \hline
    Database & Release year & Modalities & \makecell[c]{\#Images or Videos\\(Live, Spoof)} & Spoof medium & Acquisition device\\
    \hline
    ZJU EyeBlink \cite{pan2007eyeblink} & 2007 & RGB & ( 80 , 100 ) & High-quality photo & Webcam (320$\times$240)\\
    \hline
    NUAA \cite{tan2010face} & 2010 & RGB & ( 5105 , 7509 ) & A4 paper & Webcam (640$\times$480)\\
    \hline
    \makecell[c]{IDIAP Print \\Attack \cite{anjos2011counter}} & 2011 & RGB & ( 200 , 200 ) & A4 paper & MacBook Webcam (320$\times$240)\\
    \hline
    CASIA FASD \cite{zhang2012face} & 2012 & RGB & ( 200 , 450 ) & \makecell[c]{iPad 1 (1024$\times$768)\\Printed photo} & \makecell[c]{Sony NEX-5 (1280$\times$720)\\USB camera (640$\times$480)\\Webcam (640$\times$480)}\\
    \hline
    \makecell[c]{IDIAP Replay \\Attack \cite{chingovska2012effectiveness}} & 2012 & RGB & ( 200 , 1000 ) & \makecell[c]{iPad 1 (1024$\times$768)\\iPhone 3GS (480$\times$320)} & \makecell[c]{MacBook Webcam (320$\times$240)\\Cannon PowerShot\\SX 150 IS (1280$\times$720)}\\
    \hline
    3DMAD \cite{nesli2013spoofing} & 2013 & RGB, Depth & (51100, 25500) & 3D Mask & \makecell[c]{Microsoft
    Kinect for\\Xbox 360 (640$\times$480)}\\
    \hline
    \makecell[c]{MSU-MFSD \cite{wen2015face}} & 2015 & RGB & ( 110 , 330 ) & \makecell[c]{iPad Air  (2048$\times$1536)\\iPhone 5s (1136$\times$640)\\A3 paper} & \makecell[c]{Nexus 5 (720$\times$480)\\MacBook (640$\times$480)\\Canon 550D (1920$\times$1088) \\iPhone 5s (1920$\times$1080)}\\
    \hline
    MSU-RAFS \cite{patel2015live} & 2015 & RGB & ( 55 , 110 ) & MacBook (1280$\times$800) & \makecell[c]{Nexus 5 (1920$\times$1080)\\ iPhone 6 (1920$\times$1080)}\\
    \hline
    \makecell[c]{IDIAP Multi-\\spectral-Spoof \cite{chingovska2016face}} & 2016 & RGB, Near-Infrared & ( 1689 , 3024 ) &A4 paper &u-Eye camera (1280$\times$1024)\\
    \hline
    MSU-USSA \cite{patel2016secure} & 2016 & RGB & ( 1140 , 9120) & \makecell[c]{MacBook (2880$\times$1800)\\Nexus 5 (1920$\times$1080)\\Tablet (1920$\times$1200)\\11$\times$8.5 in. paper} & \makecell[c]{Nexus 5 (3264$\times$2448)\\Cameras used to\\ capture celebrity photos}\\
    \hline
    HKBU-MARs \cite{liu20163d2} & 2016 & RGB & (504, 504) & 3D Mask & \makecell[c]{Logitech C920, industrial camera,\\Canon EOS M3, Nexus 5, iPhone 6\\Samsung S7, Sony Tablet S}\\
    \hline
    OULU-NPU \cite{boulkenafet2017oulu} & 2017 & RGB & ( 1980 , 3960 ) & \makecell[c]{A3 paper\\Dell UltraSharp 1905FP\\ Display (1280$\times$1024)\\MacBook 2015 (2560$\times$1600)} & \makecell[c]{Samsung Galaxy S6 edge\\ HTC Desire EYE, OPPO N3\\MEIZU X5, ASUS Zenfore Selfie\\Sony XPERIA C5 Ultra Dual}\\
    \hline
    \makecell[c]{SiW (Spoofing\\in the Wild)\cite{liu2018learning}} & 2018 & RGB & ( 1320 , 3300 ) & \makecell[c]{Samsung Galaxy S8\\iPhone 7, iPad Pro\\ PC (Asus MB168B) screen} & \makecell[c]{Canon EOS T6\\ Logistech C920 webcam}\\
    \hline
    ROSE-YOUTU \cite{li2018unsupervised} & 2018 & RGB & 4225 & \makecell[c]{A4 paper\\Lenovo LCD (4096$\times$2160)\\Mac screen (2560$\times$1600)}&\makecell[c]{Hasee smartphone (640$\times$480)\\Huawei Smartphone (640$\times$480)\\iPad 4 (640$\times$480)\\iPhone 5s (1280$\times$720)\\ZTE smartphone (1280$\times$720)}\\
    \hline
    IDIAP-CSMAD \cite{bhattacharjee2018spoofing} & 2018 & \makecell[c]{RGB, near-infrared\\ (NIR), Thermal from \\long-wave infrared (LWIR)} & (87, 159) & 3D Mask & \makecell[c]{Realsense SR300,\\Compact Pro} \\
    \hline
    3DMA \cite{xiao20193dma} & 2019 & RGB, NIR & (536, 384) & 3D Mask &  \makecell[c]{R0710A binocular\\camera (640$\times$480)}\\
    \hline
    % CASIA-SURF CeFA\cite{liu2021casia} & 2018 & \makecell[c]{RGB, Depth,\\ Infrared (IR)} & ( - , - ) & &Intel Realsense\\
    % \hline
    \makecell[c]{CUHK MMLab\\CelebA-Spoof \cite{CelebA-Spoof}} & 2020 & RGB & 625,537 & \makecell[c]{A4 paper\\Face mask, PC} & \makecell[c]{24 sensors with 4 types\\(PC, Camera, Tablet, Phone)}\\
    \hline
    \makecell[c]{CASIA-SURF \\3D Mask \cite{yu2020fas}} & 2020 & RGB & (288, 864) & 3D Mask &  Apple, Huawei, Samsung\\
    \hline
    \makecell[c]{CASIA-SURF \\3D HiFi Mask \cite{liu2022contrastive}} & 2020 & RGB & (13650 , 40950) & 3D Mask &  \makecell[c]{iPhone11, iPhoneX, MI10,\\ P40, S20, Vivo, HJIM}\\
    \hline
    Ambient-Flash \cite{di2020rethinking} & 2021 & \makecell[c]{RGB, additional \\light flashing} & ( 7503 , 7503 ) & \makecell[c]{Printed paper\\Digital screen} &\makecell[c]{Logitech C920 HD webcam\\ LenovoT430u laptop \\ webcam  (640$\times$480),  MotoG4}\\
    \hline
    Echo-Spoof \cite{kong2022beyond} & 2022 & RGB, Acoustic & ( 82,850 ,  166,666 ) & \makecell[c]{A4 paper\\iPad Pro (2388$\times$1668)\\iPad Air 3 (2224$\times$1668)}& \makecell[c]{Samsung  Edge Note (2560$\times$1440)\\Samsung Galaxy S9 (3264$\times$2448)\\Samsung Galaxy S21(4216$\times$2371)\\Xiaomi Redmi7 (3264$\times$2448)}\\
    \hline
\end{tabular}}
\end{table*}

\subsection{Overview of presentation attack detection methodologies}
\subsubsection{Face liveness detection systems and mobile applications} Face liveness detection techniques have been widely deployed in real-world applications. With various sensors ($e.g.$, RGB camera, speaker, microphone, accelerometer, etc.) assembled, most devices such as smartphones are able to take advantage of multi-modality information captured by different sensors to conduct more accurate and generalized PAD. Thanks to the pervasive availability of speakers and microphones on mobile devices, acoustic signals have been demonstrated to be effective in capturing biometric information from users for various mobile-oriented applications. Recently, great efforts~\cite{echoPrint2021, chen2019echoface, xu2021rface, kong2022beyond} have been devoted to devising acoustic-based face liveness detection frameworks to perform more reliable PAD in practical scenarios. EchoPrint~\cite{echoPrint2021}, for the first time, incorporated both RGB vision modality and acoustic modality to conduct the user authentication. However, face anti-spoofing has been largely ignored in EchoPrint. Follow-up works such as Echoface~\cite{chen2019echoface} achieved more than 96\% accuracy by using acoustic signals solely. Rface~\cite{xu2021rface} demonstrated that radio frequency signals could identify both 2D print/replay and 3D mask attacks with high-level accuracy. Moreover, EchoFAS~\cite{kong2022beyond}, designed a more advanced signal configuration and aggregated CNN and vision transformer to achieve outstanding PAD performance. Besides acoustic signal, some recent works~\cite{chan2017face, ebihara2021efficient, tang2018face} proposed to use the flash to conduct a secure face liveness detection, based on the theory that the reflection characteristics of bonafide and PAs are distinguishable. FaceRevelio~\cite{farrukh2020facerevelio} demonstrated that varying illumination could enable reconstructing the 3D face surface of the input, thereby achieving robust and accurate face liveness detection. Given the fact that dual-pixel sensors have been widely built in mobile devices, Wu $et~al.$~\cite{wu2020single} proposed to capture dual-pixel images to reconstruct the depth map and subsequently distinguish the bonafide from PAs. Chen $et~al.$~\cite{chen2017your} used an RGB camera to conduct PAD by comparing the rPPG maps of face and fingertip videos, which should be highly consistent if they are captured from a live person. Apart from the sensors mentioned above, the accelerometer and gyroscope have also been incorporated into face liveness detection. FaceLive~\cite{li2015seeing} employed accelerometer and gyroscope to measure the movement data, and the head pose video was meanwhile recorded by the built-in camera. Then the consistency between the two modality data would be used to discriminate the bonafide from attacks.

\begin{figure}[h]
\centering
\includegraphics[scale=0.38]{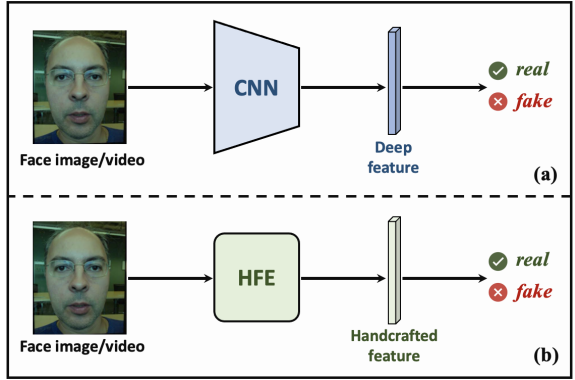}
\caption{(a) Deep learning methods for FAS. (b). Traditional handcrafted methods for FAS. HFE indicates a variety of hand-crafted feature extraction algorithms.}
\label{FAS_featrue}
\end{figure}

\begin{figure}[h]
\centering
\includegraphics[scale=0.45]{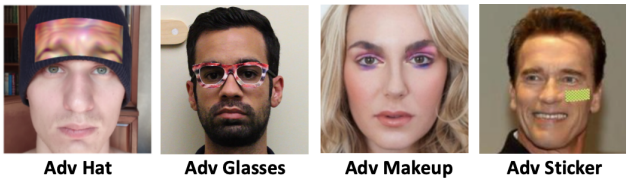}
\caption{Physical adversarial face attack samples: adv. hat~\cite{komkov2021advhat},  adv. exmples~\cite{sharif2019general}, adv. makeup~\cite{yin2021adv}, and adv. sticker~\cite{guo2021meaningful}}
\label{Adv_physical}
\end{figure}

\subsubsection{Face presentation detection methodologies} In this survey, we classify PAD methods into three categories: traditional hand-craft methods, deep learning methods, and hybrid methods. Traditional hand-craft methods attempt to extract hand-crafted features such as local binary pattern (LBP) \cite{ahonen2006face}, scale-invariant feature transform (SIFT)~\cite{patel2016secure}, histograms of oriented gradients (HoG)~\cite{dalal2005histograms}, speeded-up robust features (SURF)~\cite{boulkenafet2016face}, and difference of gaussian (DoG)~\cite{tan2010face} to perform face liveness detection. As illustrated in Fig.~\ref{FAS_featrue} (a), deep learning methods, empowered by effective neural network architectures, aim at directly extracting deep features from input face images/videos for PAD. In turn, hybrid methods assemble handcrafted feature extraction and deep feature extraction modules into one framework for final decision-making.

\noindent\textbf{Traditional handcraft methods.} We illustrate the general pipeline of handcraft method in Fig.~\ref{FAS_featrue} (b), where HFE indicates a variety of hand-crafted feature extractors such as LBP~\cite{ahonen2006face}, SIFT~\cite{patel2016secure}, HoG~\cite{dalal2005histograms}, SURF~\cite{boulkenafet2016face}, and DoG~\cite{tan2010face}. LBP~\cite{ahonen2006face} was taken as a local texture descriptor that assigned a binary label to each pixel, and the binary number was determined by the values of the central pixel and its neighbor pixels. SIFT~\cite{patel2016secure} had been widely employed to capture image representations in many computer vision tasks as it was invariant to various image distortions, such as rotation, scale, translation, etc. Besides, HoG~\cite{dalal2005histograms} showed great superiority in capturing representative features of images compared with previous edge and gradient based descriptors. SURF~\cite{boulkenafet2016face} was a fast and efficient scale and rotation invariant descriptor. It could effectively speed-up the computation and extract robust scale-independent features to perform accurate face liveness detection. Moreover, the DoG~\cite{tan2010face} filter could effectively remove the noise in the high-frequency domain, hence empowering high-performance face anti-spoofing. 

\noindent\textbf{Learning-based methods.} Thanks to the advent of deep learning, enormous progress has been achieved in this research field. Early deep learning attempts on FAS date back to 2014, where \cite{yang2014learn} first proposed to design a convolutional neural network with some data pre-processing, such as spatial and temporal augmentations, to achieve an outstanding FAS performance. Face presentation detection can be regarded as a binary classification problem. Lucena $et~al.$ \cite{lucena2017transfer} found that pretraining the VGG16~\cite{simonyan2014very} on ImageNet~\cite{russakovsky2015imagenet} and transferring the learned knowledge to FAS could effectively save computational resources and avoid the overfitting problem. With the rapid progress in network architectures, more advanced networks such as the siamese network~\cite{hao2019face} and vision transformer~\cite{george2021effectiveness} have been applied to the FAS task. Moreover, Chen $et~al.$ \cite{chen2019attention} designed a two-stream framework complementarily combining RGB feature and multi-scale retinex (MSR) feature via an attention-based fusion module and achieved outstanding generalization capability. Deb $et~al.$ \cite{deb2020look} demonstrated that local face patches could effectively reflect the inherent cues for more generalized detection. Similarly, Wang $et~al.$ \cite{wang2022patchnet} designed PatchNet to mine informative local cues and proposed asymmetric margin-based classification loss and self-supervised similarity loss to regularize the patch embedding space. On the other hand, PAD against video replay attacks plays a critical role in securing automatic face recognition systems. As such, some methods proposed to employ LSTM \cite{ge2020face, xu2015learning} and RNN \cite{muhammad2019face} to detect the temporal consistency. Yang $et~al.$ \cite{yang2019face} exploited a novel spatial-temporal network to capture subtle evidence in both spatial and temporal domains. 

Generally speaking, binary supervision can easily cause severe overfitting problems ($i.e.$, lack generalization capability to unseen environments). To mitigate the domain gap between training and testing data, many methods seek to use auxiliary supervision in the training phase, such as the binary mask \cite{liu2020disentangling, hossain2020deeppixbis, yu2020auto, george2019deep, sun2020face, liu2019deep} and depth map \cite{atoum2017face, liu2018learning, peng2020ts, wang2020deep, yu2020searching, cai2022learning, nikisins2019domain, liu2021face, george2021cross, shen2019facebagnet, yu2020multi}. Binary mask-based methods assigned 0/1 to each pixel in bonafide/fake regions. The idea of the depth map is based on the fact that live faces contain rich 3D facial structures while 2D FPA can barely reflect depth information. Typically, Sun $et~al.$ \cite{sun2020face} demonstrated that the local label supervision scheme, including local depth map and local binary label supervisions, is superior to the global binary supervision for FAS. George $et~al.$ \cite{george2019deep} conducted pixel-wise binary supervision at the feature level, thereby achieving a more accurate and robust detection performance. Liu $et~al.$ \cite{liu2019deep} designed a deep tree learning scheme with binary map supervision for zero-shot face anti-spoofing. On the other hand, depth map supervision has been widely used in FAS since it can reflect rich intrinsic spoofing cues for 2D PAD. Liu $et~al.$ \cite{liu2018learning} designed a CNN-RNN framework and simultaneously estimated depth and rPPG maps for FAS at the video level. Yu $et~al.$ \cite{yu2020searching} proposed central difference convolutional operators and extended this work by further incorporating central difference pooling \cite{yu2020fas} to estimate the depth maps for PAD. Moreover, since the reflection characterastics of PAs and bonafide are discriminative, some works~\cite{kim2019basn, zhang2020celeba, yu2020face} seek to use auxiliary geometric information such as reflection maps to conduct generalized face anti-spoofing. 
Benefiting from the advances of FAS systems, abundant auxiliary modality information is available in practical applications. For this reason, many methods propose to conduct robust PAD via multi-modal fusion. Besides RGB space, some detectors \cite{kuang2019multi, boulkenafet2015face, boulkenafet2016face} demonstrated that HSV and YCbCr could provide informative clues. Recently, researchers found that near-infrared (NIR) modality \cite{kuang2019multi, liu2021data, nikisins2019domain, liu2021face, jiang2020face, shen2019facebagnet, yu2020multi, zhang2019feathernets} contains abundant discriminative and generalized information than RGB and depth data since NIR measures the amount of heat radiated from a live face. Specifically, Liu $et~al.$ \cite{liu2021data} proposed a multi-modal two-stage cascade framework that fused three modalities of RGB, depth map, and NIR to perform PAD. \cite{liu2021face} proposed a modality translation-based FAS method that translated the RGB face image into more generalized NIR image, thereby achieving an excellent generalization capability. Besides NIR modality, rPPG signals have also been exploited in PAD since rPPG signals can reflect periodic heart rhythms of input faces. Some models~\cite{li2016generalized, yu2022benchmarking, lin2019face, liu2018learning, yu2019remote} attempt to incorporate rPPG modality to mine inherent face spoofing cues and conduct more robust FAS. 

To further improve the generalization capability of PAD, researchers recently turned to domain generalization and domain adaptation algorithms that have been demonstrated effective in a wide variety of tasks, including computer vision, natural language processing, and multi-modality problems. Domain adaptation aims at learning a model on source domain data that can adapt well to target domains with different data distributions. Recently, various domain adaptation-based methods have been proposed for FAS \cite{li2018unsupervised, wang2020unsupervised, wang2019improving, zhou2019face, mohammadi2020domain, wang2021self, jia2021unified}. Li $et~al.$ \cite{li2018unsupervised}, for the first time, used the knowledge of domain adaptation to tackle the FAS problem. The authors proposed to minimize the Maximum Mean Discrepancy (MMD) to align the distributions of training and test datasets in high-dimension feature space. Wang $et~al.$ \cite{wang2020unsupervised} designed an unsupervised adversarial domain adaptation framework to learn domain-invariant features for robust FAS. To overcome the problem that the target domain data is always unavailable in the training stage, Wang $et~al.$ \cite{wang2021self} designed a meta-learning based model that could adapt better to target domains. On the other hand, domain generalization aims at learning a robust model on source domains that can generalize well to unforeseen target domains. Due to the uncontrollable environmental variables ($e.g.,$ illumination, capture device, and attack types), the trained PAD models tend to easily suffer significant performance drops in practical applications. For this reason, extensive efforts \cite{baweja2020anomaly, jia2020single, george2020learning, nikisins2018effectiveness, perez2019deep, shao2019multi, xiong2018unknown, li2020unseen, wang2020cross, li2018domain, liu2021adaptive, chen2021camera} have been devoted to domain generalization-based FAS methods in the recent few years. Typically, Jia $et~al.$ \cite{jia2020single} proposed a single-side domain generalization model to obtain compact and generalized features on the real side. Similarly, George $et~al.$ \cite{george2020learning} designed a multi-channel CNN and used a one-class classifier to learn compact embedding for the bonafide class. Besides, Shao $et~al.$ \cite{shao2019multi} proposed to use adversarial learning to align the feature distributions between source and target domains. Wang $et~al.$ \cite{wang2020cross} used disentangled representation learning to disentangle spoofing-related features from subject-related features and achieved outstanding generalization performance. Chen $et~al.$ \cite{chen2021camera} designed a two-branch framework to capture camera-invariant features for robust PAD. Last but not least, more effective learning schemes such as zero-shot learning \cite{qin2020learning}, meta learning \cite{shao2020regularized, perez2020learning, liu2019deep, cai2022learning, wang2021self, chen2021generalizable}, knowledge distillation \cite{li2020face, li2022one}, and progressive transfer learning \cite{quan2021progressive} have been deployed to PAD and achieved promising generalization capability. 
% Domain generalization. \cite{chen2021generalizable, baweja2020anomaly, jia2020single, george2020learning, nikisins2018effectiveness, perez2019deep, shao2019multi, xiong2018unknown, li2020unseen, wang2020cross, li2018domain, liu2021adaptive} 

% Zero-shot/few-shot learning \cite{qin2020learning}, meta learning \cite{shao2020regularized, perez2020learning, liu2019deep, cai2022learning, wang2021self}
% knowledge distillation \cite{li2020face, li2022one}, progressive transfer learning \cite{quan2021progressive}

\begin{figure}[h]
\centering
\includegraphics[scale=0.32]{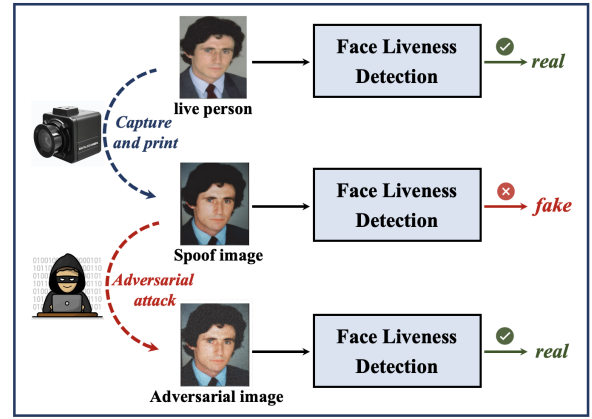}
\caption{The spoofing detection model can effectively discriminate bonafide from presentation attacks. However, expert attacker is able to apply adversarial attack technologies on spoof images to nullify the detection model.}
\label{adv_digital}
\end{figure}

\noindent\textbf{Hybrid methods.} Hybrid methods aim at taking advantage of discriminative handcrafted features and powerful learning-based models for PAD. These methods can be typically divided into three categories: \ding{182} Extracting handcraft features first and then feeding them forward to neural networks~\cite{khammari2019robust, li20203d, yu2021transrppg, li2016original}; \ding{183} Using deep models to extract deep features first and subsequently extracting handcrafted features from deep features~\cite{shao2018joint, li2019replayed, asim2017cnn}; \ding{184} Handcraft features and deep features are fused together for final classification~\cite{yu2021transrppg, rehman2019perturbing, feng2016integration, sharifi2019score, rehman2020enhancing}. To be more specific, Li $et~al.$ \cite{li2019replayed} demonstrated that motion blurs could reflect informative clues for discriminating the bonafide from replay attacks. They first extracted the motion blur indicator for each input video and then applied 1D CNN to extract deep features. Besides, TransRPPG~\cite{yu2021transrppg} designed a novel transformer-based framework for 3D mask PAD. TransRPPG first extracted the rPPG map from an input video and then designed a two-branch ViT to extract rPPG and environmental features for final decision-making. Feng $et~al.$ \cite{feng2016integration} proposed a neural network-based method by synchronously aggregating the image quality and motion cues from input face videos to conduct FAS. For the second category, Shao $et~al.$ \cite{shao2018joint} proposed deep dynamic textures for 3D mask FAS by using pre-trained VGG Net to extract deep features and then applying optical flow \cite{barron1994performance} to form the deep dynamic texture features. Li $et~al.$ \cite{li2016original} proposed to fine-tune the VGG-face model to extract deep features and then used PCA~\cite{abdi2010principal} to overcome overfitting problems. Moreover, Rehman $et~al.$ \cite{rehman2019perturbing} combined deep features and HOG maps of input face images to perform PAD. Rehman $et~al.$ \cite{rehman2020enhancing} designed a FAS framework that enhanced discriminative features by aggregating deep features and LBP texture maps of input faces. Overall, although some hybrid methods can benefit from the advantages of handcrafted features and deep features, they still have obvious drawbacks, such as needing expert prior knowledge for handcrafted feature extraction. Thus, they cannot guarantee a global optimum for FAS.

% measuring the consistency between device movement data from
% the inertial sensors and the head pose changes from the facial video
% captured by built-in camera.

\subsection{Counter-forensics issues}
The advent of effective detection tools always comes with more powerful attack methods since attacks and defenses are in an arms race. Skilled attackers are able to launch adversarial attacks to bypass PAD models. Herein, we generally categorize them into two types: physical adversarial attacks and digital adversarial attacks. 

\begin{figure}[h]
\centering
\includegraphics[scale=0.29]{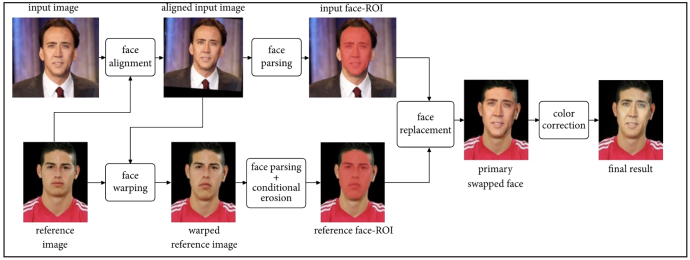}
\caption{Typical pipeline for face swapping~\cite{chen2019face}.}
\label{traditional}
\end{figure}

As shown in Fig.~\ref{Adv_physical}, physical adversarial attacks, including adversarial hat \cite{komkov2021advhat}, adversarial glasses \cite{sharif2019general}, adversarial makeup \cite{yin2021adv}, and adversarial sticker \cite{guo2021meaningful}, are capable of hacking face recognition systems under the black-box or white-box setting. Even worse, these attack methods can easily bypass most existing PAD models since the live persons are indeed actively present in front of the devices. Thus, how to design more generalized FAS models to counter the evasion of PAD remains an open problem.

Besides physical adversarial attacks, digital adversarial attacks also pose a significant challenge in this research community. As illustrated in Fig.~\ref{adv_digital}, although the existing spoofing detection model can effectively distinguish bonafide from PAs, the expert attacker can launch adversarial attacks on input media content to fool the detection model. With the rapid developments of adversarial attack algorithms, more and more attack methods \cite{zhang2020adversarial, agarwal2019deceiving,  agarwal2019deceiving2} on spoof faces have been recently proposed to deceive existing FAS models, among which \cite{zhang2020adversarial} can even achieve a 100\% attack success rate. There is no doubt that these attack methods will cause severe security concerns and hazardous crises. Therefore, it is non-trivial to develop a more secure and robust FAS model to counteract the menace of these face adversarial attack techniques.

\subsection{Future research directions} 
To date, there are still many open issues that need to be properly addressed in the FAS research field. On the one hand, industry is now somewhat ahead of academia. For instance, Apple FaceID takes advantage of the aggregation of three modules: a dot projector, a flood illuminator, and an infrared camera to capture the 2D infrared face image and reconstruct the 3D facial structure. However, in the research community, most existing FAS databases are somewhat outdated. More advanced PA databases are expected in the future. On the other hand, generalized PAD is a long-standing challenge in this research area. Mining inherent spoof clues and designing more effective networks are necessary to empower the generalization capability. In addition, the problem of privacy leakage during the face recognition process has raised pressing concerns. Proposing privacy-reserved PAD methodologies is also of great importance to address the concerns and secure the user privacy. 

\subsection{Discussion}
In this section, we comprehensively reviewed the existing literature on PAD in terms of face spoofing datasets, PAD systems, and PAD methodologies. We further analyzed the existing security issues and main risks of adversarial attacks. Moreover, we outlined promising future research areas in physical face attacks to facilitate the development of both industry and academia.  

\begin{figure}[h]
\centering
\includegraphics[scale=0.38]{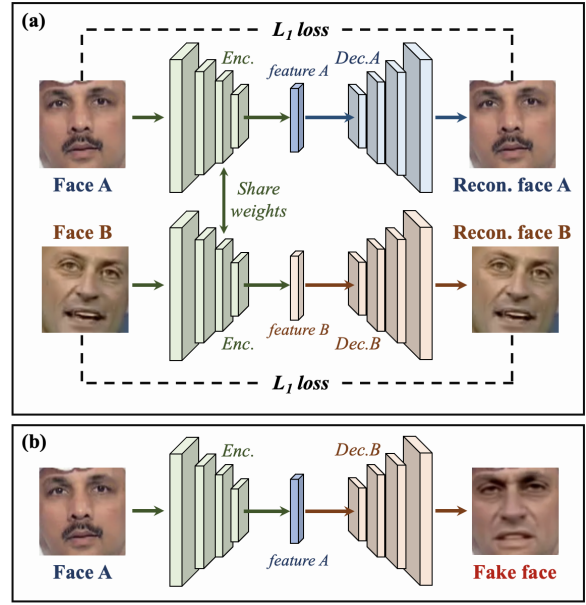}
\caption{Deepfake content creation pipeline. (a). Training phase: two face auto-encoders with one identical encoder and two specific decoders are trained under the supervision of face reconstruction loss; (b). Inference phase: feed forward the source face to the encoder and employ the target person's decoder to produce the fake face.}
\label{DF_creation}
\end{figure}
    
\section{Digital face attacks}
The digital face attack on media content are actually not a new problem. The first ever attempt at face identity swap dates back to 1860, where Abraham Lincoln's head is stitched up with the body of southern politician John Calhoun \cite{Lincoln}. 
Fig.~\ref{tree_diagram} depicts the tree diagram of literature structure on digital face forensics. Previous works can be basically classified into five categories: surveys, dataset papers, attack methodologies, detection methods, and other works. Numerous existing survey papers have reviewed prior literature on face forgery attacks and detection methodologies. However, these surveys are somewhat outdated and uninspiring to neither industry nor academia. Herein, we comprehensively review the forgery generation methodologies, deepfake datasets, existing attack detection models, and counter-forensics works. Moreover, we thoroughly analyze existing issues needed to be properly addressed and propose possible future research directions.

\begin{table*}
  \caption{Summary of digital face forgery attack databases}
  \label{Deepfake_databases}
  \centering
  \renewcommand\arraystretch{1.15}
  \scalebox{1.0}{\begin{tabular}{|c|c|c|c|c|c|c|c|}
    \hline
    Database & Release year & \makecell[c]{\#Videos\\(Real, Fake)} & \makecell[c]{\#Synthetic\\Methods} & \makecell[c]{\#Face\\Per-frame} & Subjects & \makecell[c]{Deepfake\\Audio} & \makecell[c]{Finegrained\\Labeling} \\
    \hline
    UADFV \cite{yang2019exposing} & 2018. 11 & (49, 49) & 1 & 1 & 49 & N & N\\
    \hline
    DF-TIMIT \cite{korshunov2018deepfakes} & 2018. 12 & (320, 640) & 2 & 1 & 43 & N & N\\
    \hline
    FaceForensics++ \cite{rossler2019faceforensics++} & 2019. 01 & (1,000, 4,000) & 4 & 1 & 1000 & N & N\\
    \hline
    DFD \cite{dufour2019contributing} & 2019. 09 & (363, 3,068) & 5 & 1 & 28 & N & N\\
    \hline
    DFDC \cite{dolhansky2019deepfake} & 2019. 10 & (1,131, 4,113) & 2 & \~1 & 960 & N & N\\
    \hline
    Celeb-DF \cite{li2020celeb} & 2019. 11 & (590, 5,639) & 1 & 1 & 59 & N & N\\
    \hline
    DF-Forensics-1.0 \cite{jiang2020deeperforensics} & 2020. 05 & (50,000, 10,000) & 1 & 1 & 100 & N & N\\
    \hline
    ForgeryNet \cite{he2021forgerynet} & 2021. 03 & (99,630, 121,617) & 15 & 1 & 5400+ & N & Y\\
    \hline
    FFIW \cite{zhou2021face} & 2021. 03 & (10,000, 10,000) & 3 & 3.15 & - & N & N\\
    \hline
    KoDF \cite{kwon2021kodf} & 2021. 08 & (62,166, 175,776) & 6 & 1 & 403 & N & N\\
    \hline
    FakeAVCeleb \cite{khalid2021fakeavceleb} & 2022. 12 & (500, 19,500) & 4 & 1 & 500 & Y & Y\\
    \hline
\end{tabular}}
\end{table*}

\subsection{Importance of the problem}
In recent years, falsified media content has become a vital problem on social media platforms. Faces play a central role in human communication, as a person's face can emphasize a message or even convey a message in its own right \cite{frith2009role}. However, due to the unrestricted access to enormous face media content on the network, face forgery attacks aim at manipulating pristine face images/videos have posed pressing security risks to the public at large. The situation gets even worse with the advent of AI and deep learning. The fake faces generated by deep learning methods are referred to as Deepfakes in face forensics community. Empowered by Deepfakes, the quality-level and fidelity-level of fake multimedia content have been improved so rapidly that human eyes can hardly identify the authentication. Due to the zero-barrier accessibility of the high-performance face attack resources on some open platforms ($e.g.$, Github), non-expert persons without any prior professional knowledge can readily use disclosed face forgery algorithms or APIs to create sophisticated fake content for either entertaining or malicious purposes. In this vein, these techniques could be easily fueled with targeted disinformation or misinformation and cause harmful consequences over fraud, impersonation, and rumor. For this reason, it is urgent to propose effective and robust face forgery detection methods to build digital media integrity and safeguard social platforms from face forgery attacks.

\subsection{Digital face attack methodologies}
Digital face forgery can be generally classified into four categories: identity swap, face reenactment, attribute manipulation, and entire synthesis. Fig.~\ref{digital_fake} summarizes digital face attack types and the corresponding forgery regions, where white pixels indicate the forged regions. Note that the manipulation regions of identity swap and face reenactment are provided by the official FF++\cite{rossler2019faceforensics++} dataset. Some works also define the manipulation region as the absolute difference between the pristine images and the corresponding forged ones. Generally speaking, Attribute manipulation and entire face synthesis techniques tend to bring positive impacts to human lives, while identity swap and face reenactment could cause disconcerting security problems~\cite{juefei2022countering}. For this reason, this survey mainly focuses on identity swap and facial reenactment ($a.k.a.$ expression edition).

\subsubsection{Identity swap}
The first ever work on identity swap dates back to 1860, where Abraham Lincoln's head is stitched up with the body of southern politician John Calhoun \cite{Lincoln}. A typical face swap pipeline is shown in Fig.~\ref{traditional}, which can be generally divided into four components: face alignment, face warping, face replacement, and post-processing. Heading to the era of deep learning, numerous learning-based face swap frameworks have been designed that extensively boost the quality of generated fake faces. Fig.~\ref{DF_creation} illustrates a typical fake face generation pipeline. In the training phase, two face auto-encoders that share one identical encoder are trained for the source person and the target person. The encoder learns the shared information from input faces, while two decoders are responsible for capturing the specific information for the two identities \cite{kong2021appearance}. As such, in the inference stage, the source face is firstly fed forward to the encoder and subsequently passed to the target person's decoder to produce the forged face. With the advent of powerful generative models, such as autoregressive models~\cite{van2016pixel, van2016conditional}, generative adversarial networks (GANs)~\cite{goodfellow2014generative}, and variational autoencoders (VAEs)~\cite{kingma2013auto, kingma2019introduction}, generative networks have become the main-stream deepfake generation architectures. Over the past three years, numerous terrific deepfake generation methodologies have been proposed to generate forgery face images with high-level quality and realism~\cite{gao2021information, li2020advancing, nirkin2019fsgan, lu2021live, zhu2021one, xu2022region, chen2020simswap}.  

\subsubsection{Face reenactment} 
Face reenactment is also known as the expression edition. Face2Face \cite{thies2016face2face} is one of the typical expression edition methods. It proposed a real-time face reenactment method that could reenact the target video sequence of photo-realistic quality by using a three-step solution. Follow-up works such as A2V \cite{suwajanakorn2017synthesizing} designed a cross-modal framework that was able to generate high-quality mouth texture with accurate lip sync. It employed a recurrent neural network to learn the mapping from audio features to mouth shapes, thereby could synthesize realistic speech videos. Tripathy $et~al.$ \cite{tripathy2020icface} achieved facial expression transfer with a single source and target face images by using GANs. To further improve the fidelity and quality levels of synthesized videos, many powerful frameworks had been designed in recent two years.
For example, Ha $et~al.$ \cite{ha2020marionette} designed three components: image attention block, target feature alignment, and landmark transformer to fix the identity mismatch issue between the target identity and the driver identity. Recently, 3DMM \cite{wang2021one} attempted to use a single source image and a driving video to synthesize the speech video. Hyun $et~al.$ \cite{hyun2021self} further improved the quality of face reenactment in terms of appearance consistency and motion coherency in videos. The majority of the following works focused on making generated videos look more natural and realistic. Zhang $et~al.$ \cite{zhang2021facial} demonstrated that audio not only had a high correlation with lip motion but also had a low correlation with head movement and eye-blinking. Moreover, \cite{zhang2021facial} further learned to render the head pose and eye-blinking in the synthesized videos to make them more natural.

% \subsubsection{Attribute manipulation}
% \subsubsection{Entire synthesis}

\subsection{Digital fake datasets}
Based on the release time, we summarize the existing face forgery datasets in Table.~\ref{Deepfake_databases}. According to the level of scale, quality, fidelity, and the real-world application scenarios, we generally divide these datasets into three generations. $1^{st}$ generation: UADFV \cite{yang2019exposing}, DF-TIMIT \cite{korshunov2018deepfakes}, and FaceForensics++ \cite{rossler2019faceforensics++}; $2^{nd}$ generation: DFD \cite{dolhansky2019deepfake}, DFDC \cite{dolhansky2019deepfake}, and Celeb-DF \cite{li2020celeb}; $3^{rd}$ generation: DF-Forensics-1.0 \cite{jiang2020deeperforensics}, ForgeryNet \cite{he2021forgerynet}, FFIW \cite{zhou2021face}, KoDF \cite{kwon2021kodf}, and FakeAVCeleb \cite{khalid2021fakeavceleb}. The datasets are elaborated one-by-one as follows:\\

% \subsubsection{$1^{st}$ generation digital fake datasets}\\
\noindent\textbf{UADFV} \cite{yang2019exposing} consists of 49 real videos and 49 fake videos, with 17.3k frames manipulated. The deepfake videos are produced by using generative neural networks and post-processing algorithms. 

\noindent\textbf{DF-TIMIT}  \cite{korshunov2018deepfakes} contains 320 pristine videos and 640 deepfake videos generated by faceswap-GAN~\cite{dfcode} with 32 subjects. Fake videos are equally split into high-quality(HQ) and low-quality(LQ) subsets, corresponding to the face regions with different resolutions: 128$\times$128 and 64$\times$64. Compared with UADFV, DF-TIMIT has a higher diversity and a larger scale. 

\noindent\textbf{FaceForensics++} \cite{rossler2019faceforensics++} is one of the most pervasive digital face attack datasets in the community. It composites of two forgery types: identity swap and facial reenactment, with each one containing one traditional and one deep learning-based attack, resulting in four automated face manipulation methods: Deepfakes, Face2face, FaceSwap, and NeuralTextures. FaceForensics++ covers three quality levels, and each level contains 1,000 pristine videos and 4,000 manipulated videos.  

\noindent\textbf{DFD} \cite{dufour2019contributing}. Deepfake detection dataset (DFD) was released by Google/Jigsaw in 2019. It consists of 363 real videos and 3,068 deepfake videos with 28 consented subjects in various practical scenes. 

\noindent\textbf{DFDC} \cite{dolhansky2019deepfake} includes 1,131 real videos and 4,113 fake videos, which are generated by two face-swap algorithms. DFDC is of high diversity in terms of scenes and actor characteristics.

\noindent\textbf{Celeb-DF} \cite{li2020celeb} has a higher level of quality and fidelity compared with the datasets released earlier. It collects 590 real celebrity videos from YouTube and generates 5,639 fake videos based on real videos. 

\noindent\textbf{DF-Forensics-1.0} \cite{jiang2020deeperforensics} is a large-scale deepfake dataset that contains 50,000 real videos and 10,000 forged videos generated by an end-to-end automatic face swapping model. The dataset shoot videos from 100 paid actors of various ages, skin colors, nationalities, and genders. To better imitate real-world scenarios and produce more challenging videos, extensive perturbations are applied in this dataset.   

\noindent\textbf{ForgeryNet} \cite{he2021forgerynet} builds an extremely large forgery dataset with both image- and video-level labels. ForgeryNet provides 221,247 videos shot from more than 5400 subjects, and the fake videos are generated by 15 different manipulation algorithms. It synchronously facilitates the development of four vital tasks in digital face forensics: image forgery classification, spatial forgery localization, video forgery classification, and temporal forgery localization.  

\begin{figure}[h]
\centering
\includegraphics[scale=0.35]{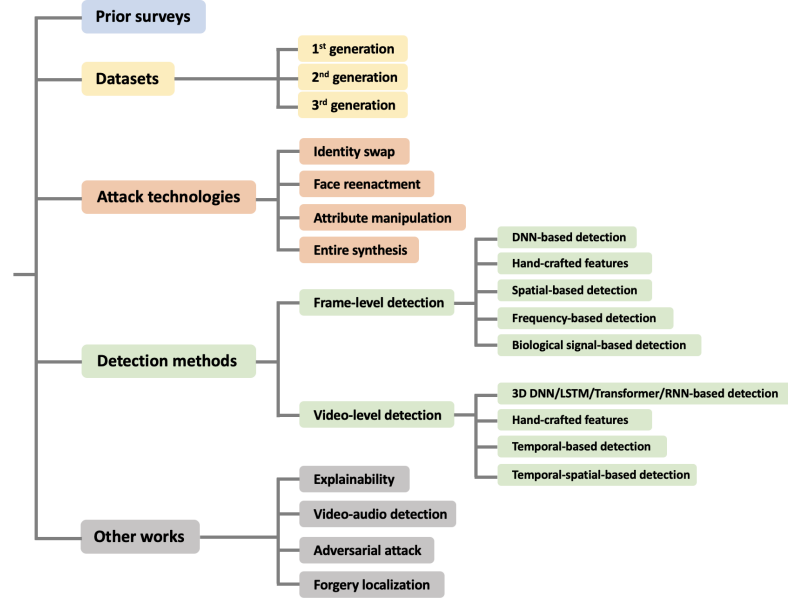}
\caption{Tree diagram of digital face attack paper structure.}
\label{tree_diagram}
\end{figure}

\noindent\textbf{FFIW} \cite{zhou2021face} constructs a large-scale and high-quality deepfake dataset by designing a novel domain-adversarial quality assessment framework. Meanwhile, it proposes a novel algorithm to tackle the multi-person problem in face forgery detection. FFIW
contains 10,000 real videos and 10,000 fake videos, with an average of more than three faces in each frame. 

\noindent\textbf{KoDF} \cite{kwon2021kodf} is a large-scale collection of deepfake and genuine videos on 403 Korean subjects. It contains 175,776 fake videos generated by six synthetic methods.

\noindent\textbf{FakeAVCeleb} \cite{khalid2021fakeavceleb} fills the gap that existing deepfake datasets either contain deepfake videos or deepfake audios. FakeAVCeleb contains 19,500 fake videos, with both videos and audios manipulated.

\subsection{Overview of face forgery detection methodologies}
Face forgery detection methodologies can be generally classified into two categories: frame-level (image-level) detection and video-level detection. The former focuses on mining key spatial or frequency information to discriminate real faces from fake ones. On the other hand, the video-level detection methods can utilize the temporal-inconsistency features to distinguish the input video clip between real and fake. We elaborate on the details of both frame-level and video-level detection as follows.

\noindent\textbf{Frame-level detection} As illustrated in Fig.~\ref{tree_diagram}, we summarize frame-level detection methodologies as the following five types: (1). DNN-based detection: DNN-based detection methods are data-driven methods, including convolutional neural networks (CNN)~\cite{afchar2018mesonet, nguyen2019capsule, chollet2017xception, tan2019efficientnet}, recurrent neural networks (RNN)~\cite{sabir2019recurrent}, and vision transformer (ViT)~\cite{heo2021deepfake}. Afchar $et~al.$ \cite{afchar2018mesonet} designed MesoNet and MesoInception4 to detect Deepfake and Face2Face videos automatically. Besides, some generic networks such as Xception Net~\cite{chollet2017xception}, Efficient Net~\cite{tan2019efficientnet}, and Capsule Net~\cite{nguyen2019capsule} have been demonstrated effective on deepfake detection tasks. Follow-up architectures such as RNN~\cite{sabir2019recurrent} and ViT~\cite{heo2021deepfake} have been employed to further improve the forgery detection accuracy. Tremendous progress has demonstrated that DNN-based methods are able to achieve promising detection methods. However, they are vulnerable to adversarial attacks and tend to suffer severe overfitting problems. (2). Hand-crafted features such as the LBP map~\cite{wang2020face}, color component~\cite{li2020identification}, and DCT map~\cite{qian2020thinking} have been taken as informative indicators for Deepfake detection; (3). Spatial-based models are the most common forgery detection methods. A wide variety of spatial-based features, such as local relations~\cite{chen2021local}, pixel region relations~\cite{shang2021prrnet}, and context discrepancies~\cite{nirkin2021deepfake}, have demonstrated their effectiveness on different datasets. Besides, some prior arts \cite{li2020sharp, zhao2021multi, kumar2020detecting} proposed to use attention mechanisms and multi-instance learning to capture informative clues in the spatial domain. (4). Frequency-based detection: F3Net~\cite{miao2022hierarchical} mined rich artifacts in the frequency domain and performed robust and accurate face manipulation detection. Miao $et~al.$~\cite{miao2022hierarchical} extended this idea by designing hierarchical frequency-assisted interactive networks to conduct more robust detection. Li $et~al.$ proposed a method that extracts frequency-aware discriminative features supervised by single-center loss. (5). Biological signal-based detection: some remote photoplethysmography (PPG) methods have been proposed to expose manipulation in synthesized videos. The basic idea of these methods is grounded on the fact that fake videos cannot replicate the biological signal of synthesized faces. In this vein, DeepRhythm~\cite{qi2020deeprhythm} utilized dual-spatial-temporal attention to capture normal heartbeat rhythms and detect deepfake videos. Similarly, \cite{ciftci2020fakecatcher} extracted ppg maps and computed spatial coherence and temporal consistency to identify the authentication of input videos.

\begin{figure}[h]
\centering
\includegraphics[scale=0.82]{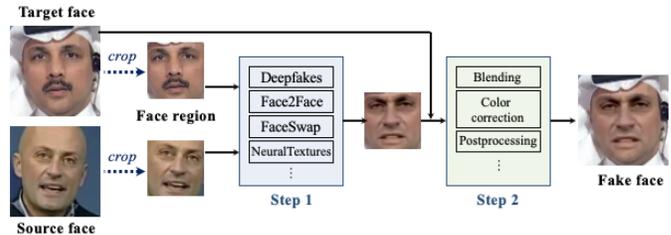}
\caption{The overview of face manipulation pipeline can be generally regarded as a two-step process. \textbf{Step 1} aims at applying various algorithms to modify the face content. \textbf{Step 2} conducts the blending, color correction, and postprocessing processes. }
\label{general_pipeline}
\end{figure}

Due to the two-player nature between face forgery and forgery detection, attack techniques are getting smarter and smarter. Previous detection methodologies can achieve outstanding detection performance under intra-settings while they are struggling in detecting unforeseen deepfake attacks or datasets. It is of great significance to mitigate these domain gaps and propose more robust and generalized detection models. As shown in Fig.~\ref{general_pipeline}, there are two general steps for generating manipulated faces. Given two input faces, \textbf{Step 1} applies face manipulation algorithms to alter the face content, and \textbf{Step 2} conducts various post processes like blending, color correction, and other post-processing. Inspired by the fake face generation pipeline, face X-ray~\cite{li2020face} focused on the blending process, which employs ground-truth boundary maps and binary labels to jointly supervise the training process. Luo $et~al.$~\cite{luo2021generalizing} found that CNN-based models tend to overfit to training data. So they proposed to use more intrinsic high-frequency noise features to conduct generalized face forgery detection. SPSL~\cite{liu2021spatial} observed that the up-sampling operation is common in most manipulation techniques, and this operation introduces unique forgery traces in the frequency domain. Shiohara $et~al.$~\cite{shiohara2022detecting} further extended this idea by incorporating more common forgery artifacts such as landmark mismatch, blending boundary, color mismatch, and frequency inconsistency to further improve the generalization capability. On the other hand, Zhao $et~al.$~\cite{zhao2021learning} hypothesized that images’ distinct source features could be preserved in manipulated faces. Based on this assumption, they proposed to measure the consistencies of image patches and achieved promising performance. Cao $et~al.$~\cite{cao2022end} built an end-to-end reconstruction architecture to learn the optimal forgery patterns. Zhu $et~al.$~\cite{zhu2021face} found that decomposing an image into several constituent elements and utilizing direct light and identity texture can remarkably extract subtle forgery patterns. Moreover, lots of powerful learning regularities such as meta learning~\cite{sun2021domain}, few-shot learning~\cite{korshunov2022improving}, contrastive learning~\cite{sun2022dual}, and neural coverage~\cite{wang2019fakespotter} have been demonstrated effective for general forgery detection. 

% Generalization: face X-ray~\cite{li2020face}, neural coverage~\cite{wang2019fakespotter}, 3D decomposition~\cite{zhu2021face}, Adversarial Example~\cite{chen2022self}, dual contrastive learning~\cite{sun2022dual}, few-shot~\cite{korshunov2022improving}, HF-features~\cite{luo2021generalizing}, learn-to-weight~\cite{sun2021domain}, pro-enhancement~\cite{gu2022exploiting}, recon.~\cite{cao2022end}, self-blended~\cite{shiohara2022detecting}, self-consistency~\cite{zhao2021learning}, spsl~\cite{liu2021spatial}\\
Besides forgery detection, forgery localization is another vital task in the face forensics community. Localizing manipulated regions of forgery faces can not only provide solid evidence for final decision-making but also unveil the potential intents of attackers. Some methods such as multi-task~\cite{nguyen2019multi}, DFFD~\cite{dang2020detection}, Detect and Locate~\cite{kong2022detect}, Ghazal $et~al.$~\cite{mazaheri2022detection}, and Fakelocator~\cite{huang2022fakelocator} have been recently proposed. They can accurately localize forgery regions and identify the authentication of input faces.

\noindent\textbf{Video-level detection}
Most video-level detection methodologies capture temporal inconsistency in fake videos and combine spatial artifacts to jointly conduct the final decision-making. Generic neural networks such as 3DCNN~\cite{zhang2021detecting}, LSTM~\cite{amerini2020exploiting, hochreiter1997long}, RNN~\cite{chintha2020recurrent, sabir2019recurrent}, and ViT~\cite{khan2021video} have achieved 
impressive detection performance. Some methodologies focus on extracting hand-crafted features like eye-blinking~\cite{li2018ictu}, head pose~\cite{yang2019exposing}, face warping~\cite{li2018exposing}, and lip movement~\cite{yang2020preventing}. Other models~\cite{sun2021improving, hu2022finfer, hu2021detecting, gu2021spatiotemporal} attempt to combine both spatial and temporal artifacts in manipulated videos and perform a more accurate deepfake detection. To defend against unforeseen attacks and datasets, more generalized and robust detectors have been designed in recent three years. DeepRhythm~\cite{qi2020deeprhythm} demonstrated that the rppg maps could reflect heartbeat rhythms, which can be further taken as a reliable and robust indicator for video-level deepfake detection. \cite{masi2020two} proposed a two-branch framework to capture the intrinsic low-level artifacts while suppressing the high-level semantic information in input videos. Haliassos $et~al.$~\cite{haliassos2022leveraging} only exploited real talking faces to conduct a more robust and generalized detection in a self-supervision manner. Lipforensics~\cite{haliassos2021lips} focused on the irregularities in mouth movement, which are common in most manipulated videos. Besides, Temporal Coherence~\cite{zheng2021exploring} proposed an end-to-end framework combining a fully connected convolution network and a temporal transformer network for extracting the temporal features and long-term temporal coherence. Moreover, some multi-modal methodologies~\cite{mittal2020emotions, zhou2021joint, kong2021appearance} jointly used visual and audio information to achieve a variety of deepfake tasks.

\subsection{Counter-forensics issues}
Although existing detectors have shown effectiveness and robustness on various face forgery datasets, they also stimulate the births of more powerful attacks. Attacks and defenses are in an arms race of such typical two-player games. Thus, it is unsurprising that adversarial attacks have recently fueled the face forensics community. Most face forgery detectors are vulnerable to both black-box and white-box adversarial attacks. Neekhara $et~al.$~\cite{neekhara2021adversarial} launched adversarial attacks on deepfake detectors in a black-box setting. They demonstrated that the designed universal adversarial perturbations could be flexibly deployed on face images and bypass forgery detectors. \cite{hussain2021adversarial} proposed that adversarial perturbations could fool DNN-based detectors and the produced adversarial videos were robust to video and image compression. Jia $et~al.$~\cite{jia2022exploring} proposed a meta-learning framework to generate more imperceptible adversarial samples by injecting adversarial perturbations into the frequency domain. Adversarial attacks have posed pressing new challenges for both industry and academia. They demand more powerful and robust face forgery detectors to properly counteract the potential risks caused by counter-forensics issues. 

\subsection{Future research directions}
Extraordinary success in deepfake attacks and digital face forensics has been achieved in the last few years. Nonetheless, there are still lots of issues that need to be addressed. Although accurate and secure, most of the deepfake detectors lack explainability and interpretability, thus limiting their reliability when deployed in practical scenarios. More explainability-related works are expected in the future to better interpret why the decision is made by the defense system, and then the decision can be adjusted accordingly. Besides accurately detecting forgery faces, localizing forgery regions is another vital task in this community. Forgery localization is able to provide evidence for detecting deepfakes and unveil attackers' intents. However, this task has been largely understudied so far. On the other hand, due to the two-player nature between face forgery and forgery detection, attack algorithms will be more and more powerful, and the generated fake faces will get increasingly realistic. This research field calls for more robust detection methods to counteract the menace of unforeseen advanced attack methods and address the generalization issues. Moreover, deepfake videos in the wild always involve both visual and audio manipulation to make the fake videos look more realistic. As shown in Table \ref{Deepfake_databases}, only FakeAVCeleb \cite{khalid2021fakeavceleb} considers deepfake audio ($a.k.a.$ audio manipulation). To facilitate accurate deepfake detection in the wild, more visual-audio joint deepfake datasets and multi-modal detectors are expected in future research works.

\begin{figure}[h]
\centering
\includegraphics[scale=0.4]{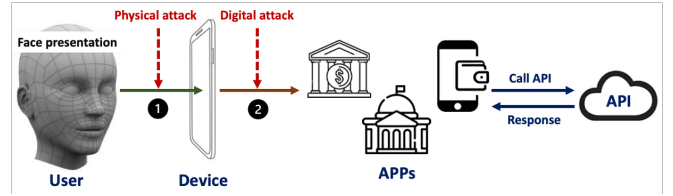}
\caption{Overview of AFR process. Step \ding{182} and step \ding{183} are vulnerable to physical and digital face attacks, respectively. }
\label{joint}
\end{figure}

\subsection{Discussion}
In this section, we comprehensively reviewed the existing digital face forgery literature over several important tasks, including face forgery generation, deepfake datasets, and face forgery detection methodologies. We thoroughly analyzed the potential risks and dangerous consequences of digital face attacks and adversarial attacks. Besides, we outlined the existing and upcoming challenges in the face forensics community and suggested possible future research directions for industry and academia. 

\section{Unifying security efforts against physical and digital face attacks}
\subsection{Importance of the problem}
Automated face recognition (AFR) systems have been pervasively deployed to billions of human beings all around the world for various applications. It is reported that the market of AFR will reach USD 3.35B by 2024~\cite{AFR_market}. However, as shown in Fig.~\ref{joint}, AFRs are vulnerable to both physical and digital face attacks. Malicious attackers can readily launch various physical attacks on the video capture stage or hack the device with digital attacks. Most current defense methods are only capable of detecting either physical or digital attacks, thereby requiring the attack type be known as a prior. Generalizing to unknown attack types has been remained as an open issue in this research community. Therefore, it is of utmost importance to propose generalized and unified detectors for safeguarding AFR systems from various malicious attacks. On the other hand, presentation attack detection and digital face forgery detection are two highly related tasks. Training a unified detector can be cast as a multi-task learning problem. \cite{yu2022benchmarking} demonstrated that the generalization capacities of models could be obviously improved via the joint training scheme compared with single-task learning. Therefore, it is of much necessary to devote more efforts to unifying security for AFR against physical and digital face attacks, which have been barely studied in the existing literature.  

\subsection{Overview of joint face spoofing and forgery detection methodologies}
Joint detection is a brand new task that requires more attention in this community. DFFD~\cite{dang2020detection} is the first attempt at unifying the detection over four digital face attacks, including identity swap, face reenactment, attribute manipulation, and entire face synthesis. On the other hand, Li $et~al.$~\cite{li2022seeing} demonstrated that face liveness verification systems are vulnerable to not only presentation attacks but also digital face attacks (Deepfake). Inspired by these two arts, follow-up works~\cite{mehta2019crafting, deb2021unified, yu2022benchmarking} attempt to propose unified detection to counteract physical face spoofing and digital face forgery. Mehta $et~al.$ \cite{mehta2019crafting} proposed to use the cross asymmetric loss function to supervise the training process and achieved promising attack detection performance in three scenarios: ubiquitous environment, individual databases, and cross-attack/cross-database.
Deb $et~al.$ \cite{deb2021unified} cast the task of unified detection of digital and physical face attacks as a multi-task problem and achieved a more generalized defense for automatic face recognition. Yu $et~al.$ \cite{yu2022benchmarking} firstly built a benchmark for joint face spoofing and forgery detection. Then they proposed a novel multi-modal framework that combined rPPG facial signals and RGB face images and achieved the best detection performance. \cite{yu2022benchmarking} demonstrated that joint training can greatly boost the generalization capability as spoofing detection and forgery detection are two highly related tasks.

\subsection{Future research directions}
By far, only few efforts have been dedicated to this unified detection task. Although the benchmark on joint physical and digital face attack detection has been built in \cite{yu2022benchmarking}, it only considered video-level detection. It is necessary to benchmark the joint detection at the image-level because image-level attacks are prominent in many real-world scenarios. Besides, standard protocols for this task should be properly built in future works to facilitate the development of new models. Apart from benchmarks and protocols, more generalized features and intrinsic clues between these two highly-related tasks are expected to be extracted to further improve the generalization capability. Last but not least, the interpretability for why the generalization capability of joint detectors boosts compared with the single-task learning scheme is still vague. More explainability and interpretability works are expected in the future.   

\subsection{Discussion}
We innovatively analyzed and discussed the pivotal joint detection problem in this section, which, to the best of our knowledge, was never mentioned in the existing surveys. The importance of this problem had been firstly clarified to attract more attention to this research field. Then, we reviewed early attempts on this joint face anti-spoofing and forgery detection task. We analyzed the main drawbacks of these works and suggested promising areas for future research. As clearly stated before, the joint detection task is largely understudied so far. To fill this gap in the face forensics community, more efforts regarding generalized unifying fake face detection are expected in future research works.    

\section{Conclusion}
Securing face data circulating on the internet and face recognition systems deployed in real-world applications is becoming a significant necessity to the public at large. Over the past decades, we have witnessed tremendous progress in both face attacks and face forensics. For sure, attack and safeguard are two players in a competitive arms race, and both of them are becoming more and more mature. Generally speaking, attack samples tend to be increasingly sophisticated and realistic, which demands powerful detection tools to counteract the pressing menace. It also requires industry and academia to design robust models to defend against various unforeseen attacks.
In this survey, we have provided a comprehensive overview and concrete discussions on the literature on both physical and digital face attacks. For each respective topic, we have provided a clear problem definition and analyzed the importance of the problem. On the other hand, the taxonomy of various attack methodologies and associated databases have been listed. We presented numerous attack detectors and analyzed their technique soundness, and also pointed out the main drawbacks of existing works. More importantly, future research directions have been highlighted in this survey for addressing unsolved problems that remained in the face forensics community. One step further, at the end of the survey, we extensively surveyed and analyzed the research works on joint face spoofing and forgery detection and concluded with suggestions for future research directions. We hope this survey can help facilitate the development of the face forensics community and attract more attention to contribute to face security.

% \section{Acknowledgement}
% This work is 

\ifCLASSOPTIONcaptionsoff
  \newpage
\fi

\bibliographystyle{IEEEtran}
\bibliography{main}

\end{document}